\documentclass[nouppercase]{kaia}
\usepackage{times}
\usepackage{amsmath}
\usepackage{algorithm}
\usepackage[switch]{lineno}
\usepackage{algorithmic}
\usepackage{balance}
\usepackage{cite}
\usepackage{amsmath,amssymb,amsfonts,amsthm}
\usepackage{textcomp}
\usepackage{xcolor}
\usepackage{hyperref}
\usepackage{url}
\usepackage{kotex}
\usepackage{todonotes}
\usepackage{subcaption}
\usepackage{booktabs}
\usepackage{multirow}
\usepackage{pgfplots}
\usepackage{nicefrac}
\usepackage[algo2e,ruled,vlined]{algorithm2e}
\usepackage{dsfont}
\usepackage{framed}
\usepackage{thmtools}
\usepackage{tabularx}
\usepackage{pgfplots}
\usepackage{graphicx} 
\usepackage{color, colortbl}

\newcommand{\sd}[1]{\textcolor{magenta}{#1}}

\newcommand{\ours}{\textsf{HSC}}

\DeclareMathOperator*{\argmax}{arg\,max}

\title{Classification of Goods Using Text Descriptions With Sentences Retrieval}

\author{Eunji Lee}{a}
\author{Sundong Kim}{b}
\author{Sihyun Kim}{a}
\author{Sungwon Park}{a}
\author{Meeyoung Cha}{b}
\author{Soyeon Jung}{c}
\author{Suyoung Yang}{c}
\author{Yeonsoo Choi}{c}
\author{Sungdae Ji}{c}
\author{Minsoo Song}{d}
\author{Heeja Kim}{d}

\affiliation{School of Computing, KAIST, \{mk35471, sihk, psw0416\}@kaist.ac.kr}{a}
\affiliation{Data Science Group, Institute for Basic Science, \{sundong, mcha\}@ibs.re.kr}{b}
\affiliation{ICT and Data Policy Bureau, Korea Customs Service, \{jsy6519, pooooh3, yschoi0817, yesican2\}@korea.kr}{c} %
\affiliation{Customs Valuation and Classification Institute, Korea Customs Service, \{csong1978, tart75\}@korea.kr}{d} %

\begin{document}

\maketitle
\begin{abstract}

The task of assigning and validating internationally accepted commodity code (HS code) to traded goods is one of the critical functions at the customs office. This decision is crucial to importers and exporters, as it determines the tariff rate. However, similar to court decisions made by judges, the task can be non-trivial even for experienced customs officers. The current paper proposes a deep learning model to assist this seemingly challenging HS code classification. Together with Korea Customs Service, we built a decision model based on KoELECTRA that suggests the most likely heading and subheadings (i.e., the first four and six digits) of the HS code. Evaluation on 129,084 past cases shows that the top-3 suggestions made by our model have an accuracy of 95.5\% in classifying 265 subheadings. This promising result implies algorithms may reduce the time and effort taken by customs officers substantially by assisting the HS code classification task.

%As the number of global transactions increases and the traded goods become complex and diversify, there is more request from importers/exporters to declare the product with the correct commodity code (HS code). Determining the HS code is an essential customs task since the tariff rate of the goods links to the code, and the decision affects the global competitiveness of the goods. Even an experienced customs officer takes several weeks to classify contentious goods and reports back to importers. The current paper addresses an HS code classification framework to facilitate the process. By understanding product description with KoELECTRA, the model suggests 6-digit HS codes (subheadings) to consider and provides key sentences and related cases to back up the decision. Tests were conducted on unknown cases collected over the last three months and showed a top-3 accuracy of 95.5\% to classify 265 subheadings.

\end{abstract}
\section{Introduction}

%According to the World Customs Organization (WCO), the number of import and export declarations worldwide reached 500 million as of 2020. In particular, as cross-border e-commerce shopping increases due to COVID-19, the number of imports and exports of e-commerce goods in Korea amounted to 63 million in the same year, which is about three times that of general import and export declarations. As the number of global transactions increases and the traded products become diversified and complicated, managing standards to categorize numerous products---Harmonized Commodity Description and Coding System (HS) is becoming crucial. HS is an international standard for classifying goods, from live animals to electronic devices, each product is classified as one of 5,387 subheadings (6-digit HS codes) according to international conventions~\cite{HS_compendium}. In addition, tariff rate, import/export requirement and availability are determined according to the HS code of the product. 

According to the World Customs Organization (WCO), the number of import and export declarations worldwide reached 500 million as of 2020. 
Events like COVID-19 have led to a surge in cross-national imports of e-commerce goods, where for instance, Korea marked 63.5 million in 2020, 48\% increase compared to the previous year~\cite{e-commerce-stats}. 
As global transactions increase and the traded products become diversified, managing standards to categorize numerous products---i.e., Harmonized Commodity Description and Coding System (HS)---becomes crucial. HS is an international standard for classifying goods; from live animals to electronic devices, each product is classified as one of 5,387 subheadings (6-digit HS codes) that meet international conventions~\cite{HS_compendium}.
This code determines critical trade decisions like tariff rate, import and export requirement, etc. 

HS code classification is non-trivial and requires a high degree of expertise since it determines the tariff rates. Securing tariffs is vital for fiscal income in many countries. The share of tax revenue secured through the customs office is nearly 20\% worldwide, and it exceeds 40\% in West African countries.\footnote{WCO annual report shows the proportion of revenue collected by customs in tax revenue of each country (pp.~46-91): \url{https://tinyurl.com/yxjvn9mz}} In addition, tariff rates directly link to the price of goods, affecting their global competitiveness. Therefore, importers and exporters pay special attention to the product declaration. Customs authorities will scrutinize the HS code of the declared goods and correct them if needed. When the HS code is wrong, the customs authorities make them correct the code. Simple errors can be corrected by amending the declaration or sending a request for correction. If customs administrations recognize the evidence of smuggling or intention of false declaration for tax evasion, importers will be punished by customs act. 

Classifying a product is complex because human experts' interpretation may not always be consistent. It can lead to international disputes when there is a difference of opinion between customs authorities or between companies and customs authorities. For example, when smartwatches were first released, tariffs varied by the importing country due to the absence of a classification standard. For example, tariff rates for wireless communication devices are 0\%, but 4--10\% for watches. This led to a dispute, which was finally resolved at the WCO HS Committee in 2014. The committee classified smartwatch as a wireless communication device, and the manufacturer was able to save about \$13 million a year~\cite{smartwatch}. Accordingly, the customs administration operates a pre-examination system, allowing import and export companies to request the customs for reviewing their items before formal declarations. There are about 6,000 applications for pre-examination every year in Korea. As the complexity of the goods increases, the processing time has increased from 20.4 days to 25.8 days since 2018. % (20.4 days in 2018 → 25.7 days in 2019 → 25.9 days in 2020). 
The main reason is the detailed review process since the HS code and the corresponding tax rate can differ even for similar-looking items. For example, tariff rates for television (HS 8528.59) are 8\%, but 0\% for PC monitor (HS 8528.52).

HS codes had been determined by reviewing the description submitted by applicants and relevant cases in the past. Experts adjust to Harmonized Commodity Description and Coding System Explanatory Notes (HS manual)~\cite{HS_manual} for standard code descriptions and General Rules for Interpretation of Nomenclature (GRI)~\cite{GRI} for decision-making criteria. In this paper, we present a novel HS classification model by reflecting on how experts work. First, the model suggests 4-digit HS codes (headings) based on product descriptions with pre-trained language models. Then, it retrieves key sentences from the HS manual that are most related to the product. Next, the model suggests 6-digit HS codes (subheadings) using product descriptions and retrieved sentences. The workflow is similar to the order of applying GRI when experts perform HS classification tasks. Retrieved sentences act as supporting facts of the decision to convince importer/exporters.

We classified headings and subheadings of recently examined electrical equipments (i.e., Chapter 85), which is known to be difficult due to product complexity. Our model outperformed the winning solution used in the product classification challenge in the e-commerce sector~\cite{lime2018}. 
Moreover, we demonstrated our model at the Korea Customs Service with eleven undisclosed decision cases (Top-3 accuracy: 0.82).
Last, we discuss how to advance our model by discussing interpretability and contextual logic understanding.

\section{Related Work}
\subsection{Harmonized System (HS) code}
All the items that go through customs are assigned the Harmonized System (HS) code, an internationally standardized system of names and numbers to classify traded products to determine tariffs. Being an internationally recognized standard, the first six digits of the HS code (HS6) are the same for all the countries. For further classification, countries have added more digits to their respective HS code systems. HS6 includes three components: 
\begin{enumerate}
\item  \textbf{Chapter}, the first 2-digit of HS code, contains 96 categories from 01 to 99. Example chapter 85 indicates electrical machinery and equipment and parts thereof. 

\item  \textbf{Heading}, the first 4-digits, groups similar characteristics of goods within a Chapter. % Heading should be examined together with Chapter. 
For instance, heading 8528 represents monitors and projectors but excludes television reception apparatus.

\item  \textbf{Subheading}, the first 6-digits, groups goods within a Heading. For example, the subheading 8528.71 includes items not designed to incorporate a video display or screen in 8528.
\end{enumerate}

\subsection{HS Code Prediction Methods.}
Recent studies utilize machine learning approaches to predict HS code using text descriptions of the declared goods. These approaches include $k$-nearest neighbor, SVM, Adaboost~\cite{DING20151462}, and neural networks~\cite{NNTHS2021}. To catch semantic information from the text, state-of-the-art studies use neural machine translator~\cite{10.1007/978-3-030-44322-1_22} and other transformer-based algorithms~\cite{luppes2019classifying}. Other studies utilized hierarchical relationships between HS codes and co-occurrence of the words by background nets~\cite{Customs2019}. There are similar studies that understand short texts and classify them in a large hierarchy using class taxonomy~\cite{shen2021taxoclass}, metadata~\cite{zhang2021match}, and hyperbolic embedding~\cite{chen2020hyperim}, that can be applied to HS prediction. However, most approaches focus on classification itself and lack of providing any explanation. 

\subsection{Sentence Retrieval.}
Sentence retrieval is commonly used for the question and answering (QA) tasks
%and machine comprehension problems with paragraphs for reasoning
~\cite{thayaparan-etal-2019-identifying, wang2019evidence}. Retrieved sentences become supporting facts and provide a detailed explanation of the answer. State-of-the-art approaches in identifying supporting factors use self-attention~\cite{wang2017gated}, bi-attention\cite{seo2017bidirectional} over paragraphs, and this is possible since most of the QA datasets with paragraphs have annotated evidence sentences themselves~\cite{yang-etal-2018-hotpotqa, Khot2020QASCAD}. In the case of the unsupervised setting, approaches such as TF-IDF~\cite{ramos2003using}, alignment-based methods~\cite{kim2017bridging} are used to find supporting sentences. Unsupervised sentence retrieval can increase the interpretability and performance in finding answers~\cite{groeneveld2020simple}.

\section{Dataset}

The study is conducted on goods belongs to electrical equipment (Chapter 85). Classifying these goods is getting trickier since electrical products are multi-functional and complex, so they do not easily fit into the existing harmonized system~\cite{park2019fallacy}. Because of its difficulties, goods belongs to Chapter 85 received the most classification requests (17.1\%, as of 2020). According to Customs Valuation and Classification Institute, it takes 37.2 days to resolve the classification requests, which is much longer than the average time required for other categories (25.9 days). Chapter 85 contains 46 headings and 265 subheadings in total. \looseness=-1

\subsection{Decision Cases}
We utilized the datasets from the customs law information portal (CLIP) from Korea Customs Service. CLIP lists customs decision cases over the past 20 years. Each case includes the item description, determined HS code, enforcement date, reasons, etc. Among them, some cases have been submitted to the council, committee, or the international committee because of the inherent ambiguities of the goods and surrounding politics. There are about 129,084 cases (122,221 international cases and 6,863 Korean cases) corresponding to Chapter 85. Among 6,863 Korean cases, 6,434 cases (93.7\%) were determined by Customs Valuation and Classification Institute\footnote{\url{https://www.customs.go.kr/cvnci/main.do}} with clear standards for decision. There are 237 contentious cases (3.5\%) that the HS council determined product HS codes. Even more trickest 192 cases (2.8\%) are resolved by the HS committee.

%  We trained our model with these decision cases to predict Harmonized System (HS) code, and we also used the HS manual to make an interpretable model.

\begin{table}[t]
\centering
\small
\caption{An example decision case including item description and corresponding HS code.}
\label{tab:decision}
\begin{tabular}{ p{0.95\linewidth} } \toprule
    \textbf{Item description}: Photovoltaic cell panel silicon (Si) embedded in plastic (EVA) and assembled a layer of glass and fiberglass and upper layer of ``Tedlar EVA", with an aluminum frame, which converts sunlight into electricity. Type cells are polycrystalline, with a maximum power of 135W. Each panel has 36 cells connected in series and the open circuit voltage is 22.1V. Incorporates type diodes ``bypass" of protection in the junction box and cables. It has no other devices that allow power directly usable. Dimensions 1008 x 992 x 35mm and a weight of 13.5kg.
    
    \medskip
    \textbf{HS code: 8541.40-9000} \\
    \bottomrule
\end{tabular}
\end{table}

\begin{table}[t]
\centering
\footnotesize
\caption{Heading-level explanation in HS manual. It provides characteristics and standards of each heading in detail, with one-liner description of every subheadings.}
\label{tab:decision}
\begin{tabular}{ p{0.95\linewidth} } \toprule
    \textbf{85.41 \quad Diodes, transistors and similar semiconductor devices;} photosensitive semiconductor devices, including photovoltaic cells whether or not assembled in modules or made up into panels; light-emitting diodes (LED); mounted piezo-electric crystals (+).
    
    \medskip
    
    \textbf{8541.40 \quad Photosensitive semiconductor devices,} including photovoltaic cells whether or not assembled in modules or made up into panels; light-emitting diodes (LED)
    
    \medskip
    
    (B) PHOTOSENSITIVE SEMICONDUCTOR DEVICES
    \medskip
    
    This group comprises photosensitive semiconductor devices in which the action of visible rays, infra red rays or ultra violet rays causes variations in resistivity or generates an electromotive force, by the internal photoelectric effect.
    
    \medskip
    
    Photoemissive tubes (photoemissive cells) the operation of which is based on the external photoelectric effect (photoemission), belong to heading 85.40. \\
    \bottomrule
\end{tabular}
\end{table}

% \begin{figure}[t]
% \centerline{
%       \includegraphics[width=\linewidth]{figures/data_example.png}}
%       \caption{Illustration of a decision case including item description and corresponding HS code. Explanations of item's Chapter, Heading and Subheading are also given.
%     %   \mc{Make it shorter by half?}
%       } 
% \label{fig:model}
% \end{figure}

% \begin{figure}[t]
% \centerline{
%       \includegraphics[width=\linewidth]{figures/manual_example.png}}
%       \caption{An example of the heading-level explanation in HS manual. Manual provides characteristics and standards of each heading (8541 here) in detail, and it includes one-liner description of every subheadings.
%     %   \mc{Make it shorter by half? Maybe combine with Fig 1?}
%       } 
% \label{fig:model}
% \end{figure}

\subsection{HS Manual}
Experts decide HS code according to Harmonized Commodity Description and Coding System Explanatory Notes (HS manual), a common set of laws worldwide. The HS manual explains each code in section, chapter, and heading level in detail. We utilized the heading-level manual to provide supporting facts of the model decision.
\begin{figure*}[t]
\centerline{
      \includegraphics[width=\linewidth]{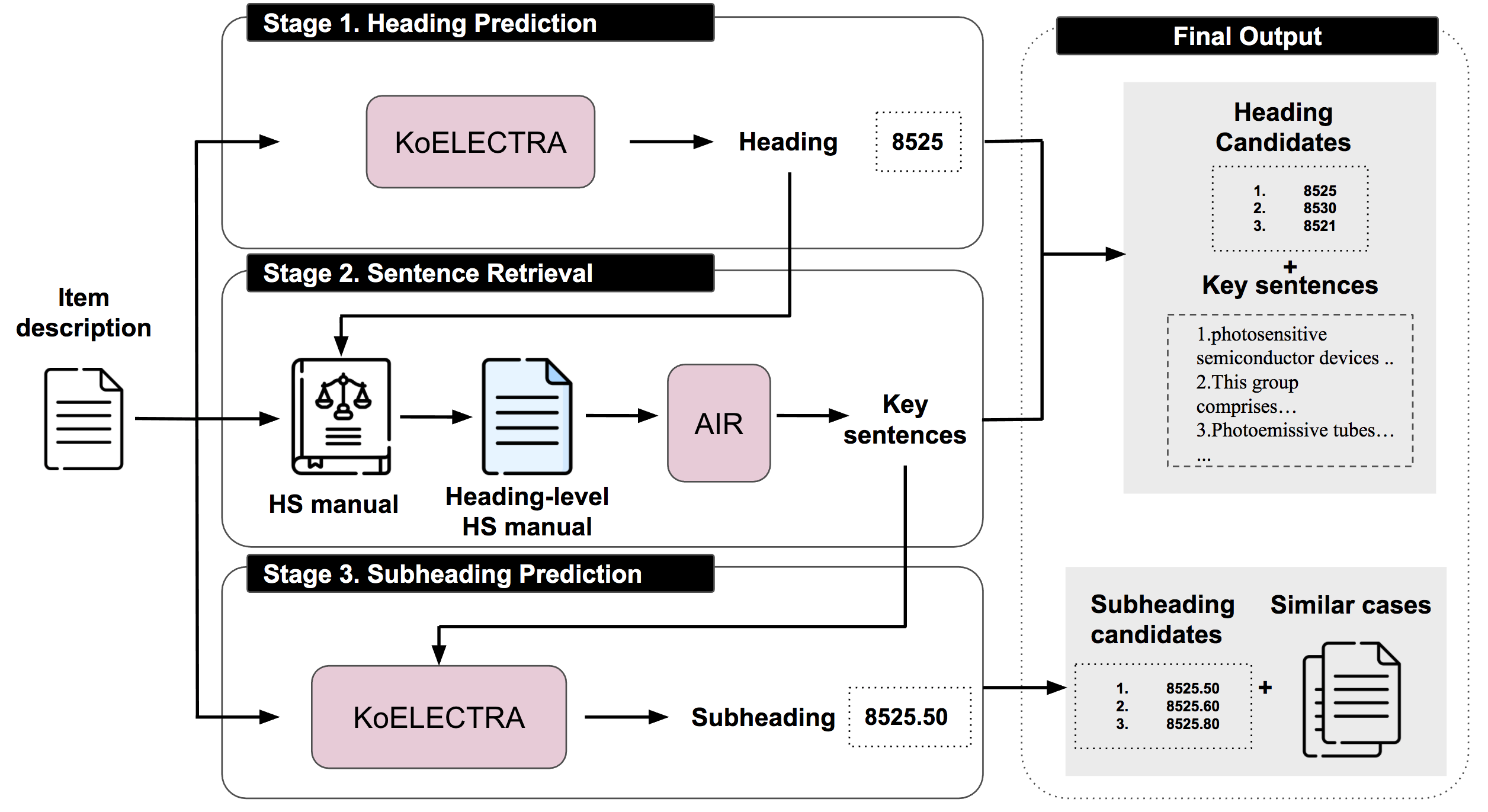}}
      \caption{The illustration of the HS code prediction system. Stage 1 uses the KoELECTRA model and predicts the heading of the product. Using the predicted heading, Stage 2 retrieves key sentences from the heading-level HS manual for supporting decisions. In stage 3, item descriptions and key sentences are utilized to predict the most relevant subheadings of the product. The final outputs include heading and subheading candidates with key sentences and similar cases for reference.
      } 
\label{fig:model}

\end{figure*}

\section{Method}
The proposed framework takes item description as an input, and the final goal is to predict the subheading (HS6) of a given item by referring to the HS manual. It also provides intermediate outputs for candidate headings and subheadings, prior cases, and key sentences from the HS manual. Figure.~\ref{fig:model} illustrates our model, which is divided into three stages. 

\subsection{Heading Prediction}
$\mathcal{D} = \{D_1,\cdots ,D_N\}$ is a collection of decision cases where each case $D_i \in \mathcal{D}$ is a pair of the item description $\mathbf{x}_{i}$ and its one-hot encoded heading label $\mathbf{y}_{i}$.
After translating all item descriptions into Korean, we used KoELECTRA~\cite{park2020koelectra} as a description encoder $e_{\theta}$ to map a sequence of words $\mathbf{x}_{i}$ into embedding space $\mathbb{R}^d$. The item embedding $e_{\theta}(\mathbf{x}_{i})$ goes through the classification head and the prediction is done by minimizing the loss $\mathcal{L}$ between the true probability $\mathbf{y}_{i}$ and the predicted probability $\mathbf{\hat{y}}_{i} = e_{\theta}(\mathbf{x}_{i})\cdot W$ where $W \in \mathbb{R}^{d\times\dim(\textbf{y}_i)}$ is a trainable weight matrix of the classification head. Since the problem is multi-class classification, categorical cross-entropy loss $H$ is used.   
\begin{align} 
    \mathcal{L} = -\frac{1}{|\mathcal{D}|}\sum_{{\mathbf{x}_i, \mathbf{y}_i}\in\mathcal{D}}  H(\mathbf{y}_i, \mathbf{\hat{y}}_i)
    \label{eq:heading_prediction}
\end{align}

\subsection{Sentence Retrieval}
After predicting the heading, the model extracts key sentences from the HS manual to justify this decision. Key sentences are iteratively retrieved by calculating the similarity score between the item description $\mathbf{x}_{i}$ and the heading-level HS manual $M$. This method is called Alignment Information Retrieval (AIR), and the process terminates when retrieved sentences cover all keywords from input descriptions from $M$ or no new keywords are discovered from $M$~\cite{yadav2020air}. 
\begin{align} 
    & s(\mathbf{x}_{i},M) = \sum_{m=1}^{|\mathbf{x}_{i}|} idf(d_m) \cdot align(d_m, M), \\
    & align(d_m, M) = \max_{k=1}^{|M|} \text{sim}(d_m, m_k), \\
    \label{eq:sentence_retrieval}
\end{align}
where $d_m$ and $m_k$ are the $m^{th}$ and $k^{th}$ sentences of the $\mathbf{x}_{i}$ and M. The cosine similarity (sim) is derived by GloVe embedding of the two inputs and idf is the inverse document frequency value. We chose maximum seven sentences from the heading-level HS manual and created a sentence set $S_i = \{{m_k}\}$ with the highest alignment score. 

\subsection{Subheading Prediction}
Using key sentences and item description, the model predicts the subheading candidates and retrieve prior cases belong to each subheading. After concatenating item description $x_i$ and key sentences $S_i$, we use KoELECTRA encoder $e_{\phi}$ to generate the embedding $e_{\phi}([\mathbf{x}_{i}, S_i])$. As we did in heading prediction, the embedding goes through the classification head and the prediction is done by minimizing the categorical cross-entropy loss between the true probability $\mathbf{y}_{i}^s$ and the predicted probability $\mathbf{\hat{y}}_{i}^s = e_{\phi}(\mathbf{x}_{i})\cdot W^s$, where $W^s \in \mathbb{R}^{d\times\dim(\textbf{y}_i^s)}$ is a trainable weight matrix of the classification head.

After training, the model can determine top-$k$ subheading candidates from the classification head. If the subheading of a given item $i$ is predicted as $y_p$, similar cases are derived in $\mathcal{D}_p \subset \mathcal{D}$, which is a set of cases whose subheading label is determined as $y_p$. Similar cases are iteratively chosen by measuring cosine similarity between item embeddings:
\begin{align} 
    \argmax_{j \in \mathcal{D}_p} \text{sim}(e_\phi(\mathbf{x}_i), e_\phi(\mathbf{x}_j)).
\label{eq:similar_cases}
\end{align}

\section{Results}
This section tests feasibility of the proposed HS classification model in terms of classification performance and retrieved sentence quality. Classification performance is measured by checking the subheading (HS6) accuracy, our final output, and heading (HS4) accuracy, the intermediate output. The quality of the retrieved sentences is examined by comparing our results with documents written by experts. \looseness=-1

\subsection{Experimental Setting}

\subsubsection{Data Preprocessing. \quad} 
For experiments, we retained 126,000 cases that the product HS code is maintained among 129,084 cases. Given that decisions may change over time, we used the last three months of data (1,466 international and 186 Korean cases) for evaluation. Three months' worth of data before the test period was used as the validation set (1,733 international cases and 102 Korean cases) for hyperparameter tuning. \looseness=-1

\subsubsection{Evaluation Metric. \quad} 
We evaluated the heading and subheading classification performance by measuring top-$k$ accuracy with $k = 1, 3, 5$. In retrieved sentences case study analysis, we measured recall and precision to evaluate the quality of the supporting factors.

\subsubsection{Training Details. \quad}
KoELECTRA models and classification heads in heading and subheading prediction are trained for 50 epochs and evaluated when validation accuracy is the highest. The embedding size of KoELECTRAs is set as 768. Key sentences of each item are required to train a subheading prediction model, so we prepared them from the answer heading's HS manual beforehand. In evaluation stage, key sentences are retrieved from the predicted heading's HS manual. Training KoELECTRA model takes 40 hours and data preparation for the sentence retrieval model takes 50 hours using NVIDIA TITAN Xp. Inference and retrieval take less than 30 seconds.

\subsection{Performance Evaluation}

\subsubsection{HS Code Prediction. \quad}
We compared the proposed model with two baselines---a word-matching model and an LSTM-based model. The word matching model computes the word matching rate between the item description and heading-level HS manual and chooses the heading with the highest matching rate. LSTM-based model~\cite{lime2018} is a winning model of the competition~\cite{kakaocompetition}, which has a similar setting with our problem: predict the detailed category of the e-commerced products using their descriptions. The model utilizes LSTM networks to get embedding from tokenized input texts. Table \ref{table:main_results} shows the top-$k$ accuracy of the baselines and our model. Two variants of our model are tested, the one used retrieved sentences from the second stage and the other did not use them.
\begin{table}[h]
\caption{HS code classification accuracy in heading (HS4) and subheading level (HS6). }
\label{table:main_results}
\centering
\resizebox{\linewidth}{!}{
\begin{tabular}{l|c|ccc}\toprule
 & HS4 & \multicolumn{3}{c}{HS6}\\
\cline{2-5}\rule{0pt}{2.5ex}Models / Top-$k$ accuracy  & $k=1$ & $k=1$ & $k=3$ & $k=5$ \\
\midrule
Word matching & 0.01 &  &   &   \\
LSTM-based & 0.687 & 0.431 & 0.673 & 0.747 \\
Ours (w/o sentences) & \textbf{0.936} & 0.891 & \textbf{0.955} & \textbf{0.968} \\
Ours & \textbf{0.936} & \textbf{0.896} & 0.937 & 0.944  \\
\bottomrule
\end{tabular}}
\end{table}

\subsubsection{Retrieved Key Sentences. \quad}
Given the item description, experts present supporting reasons for their final decision. They quote some sentences from the HS manual and provides evidence of their decision with detailed explanations. We compared the quoted sentences written by experts, with key sentences retrieved by our model. As shown in Table~\ref{tab:supporting_facts}, the key sentences (supporting facts) retrieved by our model broadly match the quotations from experts.

\begin{table}[h!]
\centering
% \scriptsize
\small
\caption{Comparison between the actual reasons for decision and supporting facts found by our algorithm.}
\label{tab:supporting_facts}
\scalebox{1}{%
\begin{tabular}{ l } \toprule
    \textbf{Reasons for decision by experts} \\
         1. Transmission apparatus ... and video camera recorders. \\
         2. TELEVISION CAMERAS, DIGITAL CAMERAS AND ... \\
         3. This group covers cameras that capture images ... \\
         4. In digital cameras and video camera recorders, ... \\ \midrule
    \textbf{Supporting facts found by our model} \\
         1. PARTS  \\
         2. TELEVISION CAMERAS, ... $\rightarrow$ Eqv. to (2) \\
         3. Transmission apparatus  ...  $\rightarrow$ Eqv. to (1) \\
         4. In digital cameras ... $\rightarrow$ Eqv. to (4) \\ \midrule
    \textbf{recall = 0.75} \\
    \textbf{precision = 0.75} \\
         
         \bottomrule
\end{tabular}
}
\end{table}

\begin{figure*}[h]
\centerline{
      \includegraphics[width=1\linewidth]{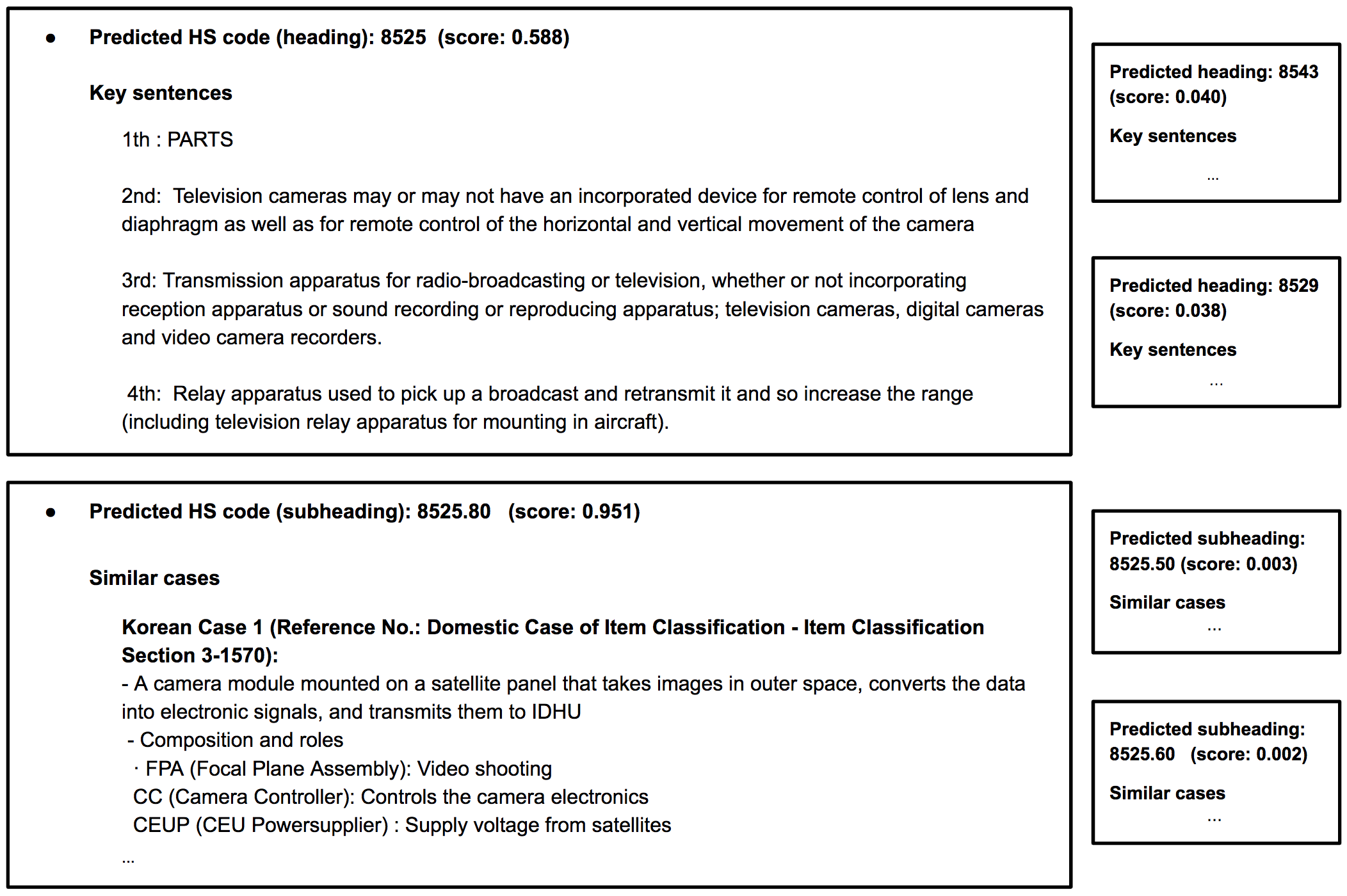}}
      \caption{Format of the model's final output. As the final result, the model provides heading and subheading candidates along with key sentences and similar cases as a basis for judgment.
      } 
\label{fig:output}
\end{figure*}

\subsection{Output Templates}
Figure \ref{fig:output} shows the model's final outputs. First, the model provides three heading candidates with their prediction scores. To calibrate the score, we applied temperature scaling to the model's softmax outputs~\cite{guo2017calibration}. For each candidate, one to seven key sentences are retrieved from the HS manual. Retrieved sentences explain the model prediction and reduce the reviewing scope by customs officers. Next, the model provides three subheading candidates. With each subheading, the model provides similar prior cases for reference. \looseness=-1

\section{Discussion}

\subsection{Towards Interpretable Results}

One of our goal is to add interpretability to the HS classification model. To do that, we provide a confidence score for each HS code candidate. The score provides additional information to judge whether a candidate is valid or not. Although the score is tuned by temperature scaling, the range of the top-$k$ confidence score was quite different for each input item. Careful calibration is required to make customs officers use this value as a reference in the decision-making process.
% \eunji{Since the score is trained by a mechanism that maximizes the value for the answer HS code, quantitative value here cannot be said as a numerical score.} 
Another way to increase the interpretability is to visualize the part of the item description relates to each subheading candidate. If then, customs officers can concentrate on the selected part and decide whether to consider second and third candidates to review. In addition, key sentences should relate to the subheading characteristics so that the final form of the model output should resemble reports written by experts. Making an organized document that explains the relation between prediction, description, and HS manual will reduce the effort required for HS code classification.

\subsection{Contextual Information in Decision}
Customs experts decide the HS code by following the General Rules for Interpretation of Nomenclature (GRI), similar to judges deciding based on the law~\cite{winter2021judicial}. On the other hand, deep learning models solve classification problems by finding common patterns from previous cases. As a result, past examples are the primary determinant of the AI model's decision, different from human experts who make their decision by rules and manuals. Since HS code and its manual undergo revision every five years, previous cases cannot always be good references for recent ones. Therefore, it is essential to employ GRIs and HS manual to make a credible model, which utilizes contextual information in model training based on deep linguistic understanding~\cite{wiegreffe2021explainableNLP}.

\section{Conclusion}
This study introduces a framework to predict harmonized system codes in customs. Using product description and HS manual, it predicts heading as subheadings and provides supporting facts and related cases to facilitate decision process by human experts. We expect that our work will contribute significantly in various aspects. This framework by declarants will improve the initial declaration quality, thereby reducing customs officials' workloads. Internally, the framework can be used to assist customs officials in carrying out their duties to increase their work efficiency and train their employees. The situation with competing HS codes is particularly problematic for declarants and customs officials. Our model presents the competing HS codes of the target product with its rationale, so it has great significance as an auxiliary means for product classification. The platforms require a systematic classification system to effectively expose and recommend products to users, but there are often different standards for each product-providing company. The platforms build and utilize hierarchical classification algorithms to maintain the consistent categorization of hundreds of millions of products. Our work will be used to advance those and facilitate their management. 

\section*{Acknowledgment}
This work was supported by the Institute for Basic Science (IBS-R029-C2, IBS-R029-Y4). We thank numerous officers from Korea Customs Service for their insightful discussions.

\bibliographystyle{plain}
% \balance
\bibliography{references}

\begin{thebibliography}{10}

\bibitem{10.1007/978-3-030-44322-1_22}
Fatma Altaheri and Khaled Shaalan.
\newblock Exploring machine learning models to predict harmonized system code.
\newblock In {\em Proc. of the European, Mediterranean, and Middle Eastern
  Conference on Information Systems}, pages 291--303, 2020.

\bibitem{kakaocompetition}
Kakao Arena.
\newblock {Product categorization competition in Daum Shopping}.
\newblock \url{https://arena.kakao.com/c/1}, 2018.
\newblock Accessed: 2021-10-07.

\bibitem{chen2020hyperim}
Boli Chen, Xin Huang, Lin Xiao, Zixin Cai, and Liping Jing.
\newblock Hyperbolic interaction model for hierarchical multi-label
  classification.
\newblock In {\em Proc. of the Thirty-Fourth AAAI Conference on Artificial
  Intelligence}, pages 7496--7503, 2020.

\bibitem{NNTHS2021}
Xi~Chen, Stefano Bromuri, and Marko van Eekelen.
\newblock Neural machine translation for harmonized system codes prediction.
\newblock In {\em Proc. of the International Conference on Machine Learning
  Technologies}, pages 158--163, 2021.

\bibitem{HS_manual}
{Ciel HS}.
\newblock Harmonized commodity description and coding system explanatory notes.
\newblock \url{http://www.clhs.co.kr/uploads/lawfile/404n3801.pdf}.
\newblock Accessed: 2021-10-07.

\bibitem{DING20151462}
Liya Ding, ZhenZhen Fan, and DongLiang Chen.
\newblock Auto-categorization of hs code using background net approach.
\newblock {\em Procedia Computer Science}, 60:1462--1471, 2015.

\bibitem{groeneveld2020simple}
Dirk Groeneveld, Tushar Khot, Mausam, and Ashish Sabharwal.
\newblock A simple yet strong pipeline for hotpotqa.
\newblock In {\em Proc. of the 2020 Conference on Empirical Methods in Natural
  Language Processing}, pages 8839--8845, 2020.

\bibitem{guo2017calibration}
Chuan Guo, Geoff Pleiss, Yu~Sun, and Kilian~Q. Weinberger.
\newblock On calibration of modern neural networks.
\newblock In {\em Proc. of the 34th International Conference on Machine
  Learning}, pages 1321--1330, 2017.

\bibitem{Khot2020QASCAD}
Tushar Khot, Peter Clark, Michal Guerquin, Peter~A. Jansen, and Ashish
  Sabharwal.
\newblock {QASC}: A dataset for question answering via sentence composition.
\newblock In {\em Proc. of the Thirty-Fourth AAAI Conference on Artificial
  Intelligence}, pages 8082--8090, 2020.

\bibitem{kim2017bridging}
Sun Kim, Nicolas Fiorini, W~John Wilbur, and Zhiyong Lu.
\newblock Bridging the gap: Incorporating a semantic similarity measure for
  effectively mapping pubmed queries to documents.
\newblock {\em Journal of biomedical informatics}, 75:122--127, 2017.

\bibitem{Customs2019}
Guo Li and Na~Li.
\newblock {Customs classification for cross-border e-commerce based on
  text-image adaptive convolutional neural network}.
\newblock {\em Electronic Commerce Research}, 19(4):779--800, December 2019.

\bibitem{luppes2019classifying}
Jeffrey Luppes, Arjen~P de~Vries, and Faegheh Hasibi.
\newblock Classifying short text for the harmonized system with convolutional
  neural networks.
\newblock {\em Radboud University}, 2019.

\bibitem{park2020koelectra}
Jangwon Park.
\newblock {KoELECTRA: Pretrained ELECTRA model for Korean}.
\newblock \url{https://github.com/monologg/KoELECTRA}, 2020.
\newblock Accessed: 2021-10-07.

\bibitem{park2019fallacy}
Minkyu Park.
\newblock {A study on the customs classification fallacy of certain ITA goods}.
\newblock {\em Korea Trade Review}, 44(2):189--202, 2019.

\bibitem{ramos2003using}
Juan Ramos.
\newblock {Using TF-IDF to determine word relevance in document queries}.
\newblock In {\em Proc. of the First Instructional Conference on Machine
  Learning}, volume 242, pages 29--48, 2003.

\bibitem{lime2018}
Lime robot.
\newblock {1st place solution of the product categorization competition in Daum
  Shopping}.
\newblock
  \url{https://github.com/lime-robot/product-categories-classification}, 2018.
\newblock Accessed: 2021-10-07.

\bibitem{seo2017bidirectional}
Minjoon Seo, Aniruddha Kembhavi, Ali Farhadi, and Hannaneh Hajishirzi.
\newblock Bidirectional attention flow for machine comprehension.
\newblock In {\em Proc. of the 5th International Conference on Learning
  Representations}, 2017.

\bibitem{e-commerce-stats}
Korea~Customs Service.
\newblock {E-commerce goods import trend}.
\newblock \url{https://tinyurl.com/4jdch8c5}, 2018.
\newblock Accessed: 2021-10-07.

\bibitem{shen2021taxoclass}
Jiaming Shen, Wenda Qiu, Yu~Meng, Jingbo Shang, Xiang Ren, and Jiawei Han.
\newblock {TaxoClass}: Hierarchical multi-label text classification using only
  class names.
\newblock In {\em Proc. of the 2021 Conference of the North American Chapter of
  the Association for Computational Linguistics: Human Language Technologies},
  pages 4239--4249, 2021.

\bibitem{thayaparan-etal-2019-identifying}
Mokanarangan Thayaparan, Marco Valentino, Viktor Schlegel, and Andr{\'e}
  Freitas.
\newblock Identifying supporting facts for multi-hop question answering with
  document graph networks.
\newblock In {\em Proc. of the Thirteenth Workshop on Graph-Based Methods for
  Natural Language Processing (TextGraphs-13)}, pages 42--51, 2019.

\bibitem{smartwatch}
{The Korea Times}.
\newblock {Smartwatch is a communication device}.
\newblock \url{https://tinyurl.com/4vrfx7ef}.
\newblock Accessed: 2021-10-07.

\bibitem{wang2019evidence}
Hai Wang, Dian Yu, Kai Sun, Jianshu Chen, Dong Yu, David McAllester, and Dan
  Roth.
\newblock Evidence sentence extraction for machine reading comprehension.
\newblock In {\em Proc. of the 23rd Conference on Computational Natural
  Language Learning}, pages 696--707, 2019.

\bibitem{wang2017gated}
Wenhui Wang, Nan Yang, Furu Wei, Baobao Chang, and Ming Zhou.
\newblock Gated self-matching networks for reading comprehension and question
  answering.
\newblock In {\em Proc. of the 55th Annual Meeting of the Association for
  Computational Linguistics (Volume 1: Long Papers)}, pages 189--198, 2017.

\bibitem{wiegreffe2021explainableNLP}
Sarah Wiegreffe and Ana Marasovi\'c.
\newblock {Teach me to explain: A review of datasets for explainable NLP}.
\newblock {\em arXiv preprint arXiv:2102.12060}, 2021.

\bibitem{GRI}
{Wikipedia}.
\newblock General rules for the interpretation of the harmonized system.
\newblock
  \url{https://en.wikipedia.org/wiki/General_Rules_for_the_Interpretation_of_the_Harmonized_System}.
\newblock Accessed: 2021-10-07.

\bibitem{winter2021judicial}
Christoph Winter.
\newblock The challenges of artificial judicial decision making for liberal
  democracy.
\newblock In {\em Judicial Decision-making: Integrating Empirical and
  Theoretical Perspectives}, forthcoming.

\bibitem{HS_compendium}
{World Customs Organization}.
\newblock {HS compendium -- The harmonized system, a universal language for
  international trade}.
\newblock
  \url{http://www.wcoomd.org/-/media/wco/public/global/pdf/topics/nomenclature/activities-and-programmes/30-years-hs/hs-compendium.pdf}.
\newblock Accessed: 2021-10-07.

\bibitem{yadav2020air}
Vikas Yadav, Steven Bethard, and Mihai Surdeanu.
\newblock Unsupervised alignment-based iterative evidence retrieval for
  multi-hop question answering.
\newblock In {\em Proc. of the 58th Annual Meeting of the Association for
  Computational Linguistics}, pages 4514--4525, 2020.

\bibitem{yang-etal-2018-hotpotqa}
Zhilin Yang, Peng Qi, Saizheng Zhang, Yoshua Bengio, William Cohen, Ruslan
  Salakhutdinov, and Christopher~D. Manning.
\newblock {H}otpot{QA}: A dataset for diverse, explainable multi-hop question
  answering.
\newblock In {\em Proc. of the 2018 Conference on Empirical Methods in Natural
  Language Processing}, pages 2369--2380, 2018.

\bibitem{zhang2021match}
Yu~Zhang, Zhihong Shen, Yuxiao Dong, Kuansan Wang, and Jiawei Han.
\newblock {MATCH}: Metadata-aware text classification in a large hierarchy.
\newblock In {\em Proc. of the Web Conference 2021}, pages 3246--3257, 2021.

\end{thebibliography}

% \newpage
% \newpage

% \onecolumn
% 	\begin{center}
% 		\section*{Summary of this paper}	
% 		\setcounter{subsection}{0}
% 	\end{center}
% \noindent\fbox{
%     \parbox{\textwidth}{
%     	\begin{quote}
% 		\subsection{Problem Setup}
% 		To predict the Harmonized System code (HS code) of the declared goods by understanding item descriptions with interpretability. 
% 		\vspace{0.5\baselineskip}
% 		\subsection{Novelty}
% 		The proposed model provides three HS6 (subheading) code candidates with 95.5\% accuracy. Furthermore, the model's final output contains the key sentences from the HS manual and previous similar cases, which is informative for customs officials.
% 		\vspace{0.5\baselineskip}
% 		\subsection{Algorithms}
% 		The proposed mechanism consists of three stages. The first stage predicts the HS4 (heading) code with KoELECTRA using item description. The second stage retrieves the key sentences from the heading-level HS manual using item description and predicted heading in stage 1. The final stage predicts the HS6 (subheading) code with KoELECTRA using item description and key sentences. 

% 		\vspace{0.5\baselineskip}
% 		\subsection{Experiments}
% 		We tested the feasibility of the proposed model's classification accuracy and interpretability. Our model predicts HS4 (heading) code with 93.6\% accuracy and predicts HS6 (subheading) code with  89.6\% accuracy. Recall and precision of the retrieved sentences were 0.75 when compared with the reasons for the decision made by customs experts.
% 		\vspace{0.5\baselineskip}
% 		\end{quote}
%     }
% }
% \thispagestyle{empty}
\end{document}